\pdfoutput=1

\documentclass[11pt]{article}

\usepackage{acl}

\usepackage{times}
\usepackage{latexsym}

\usepackage[T1]{fontenc}

\usepackage[utf8]{inputenc}

\usepackage{microtype}

\usepackage{inconsolata}

\usepackage{graphicx}

\usepackage{multirow}
\usepackage{makecell}
\usepackage{tcolorbox}
\usepackage{graphicx}
\usepackage{algorithm}
\usepackage{algorithmic}
\usepackage{amsmath}
\usepackage{amssymb}
\usepackage{CJKutf8}

\title{TAT-R1: Terminology-Aware Translation with Reinforcement Learning and Word Alignment}

\author{Zheng Li, Mao Zheng, Mingyang Song, Wenjie Yang\\
	Tencent Hunyuan\\
	\texttt{jasonzli@tencent.com}\\
}

\begin{document}
\maketitle
\begin{abstract}
Recently, deep reasoning large language models(LLMs) like DeepSeek-R1 have made significant progress in tasks such as mathematics and coding. Inspired by this, several studies have employed reinforcement learning(RL) to enhance models' deep reasoning capabilities and improve machine translation(MT) quality. However, the terminology translation, an essential task in MT, remains unexplored in deep reasoning LLMs. In this paper, we propose \textbf{TAT-R1}, a terminology-aware translation model trained with reinforcement learning and word alignment. Specifically, we first extract the keyword translation pairs using a word alignment model. Then we carefully design three types of rule-based alignment rewards with the extracted alignment relationships. With those alignment rewards, the RL-trained translation model can learn to focus on the accurate translation of key information, including terminology in the source text. Experimental results show the effectiveness of TAT-R1. Our model significantly improves terminology translation accuracy compared to the baseline models while maintaining comparable performance on general translation tasks. In addition, we conduct detailed ablation studies of the DeepSeek-R1-like training paradigm for machine translation and reveal several key findings. The code, data, and models will be publicly released\footnote{\url{https://github.com/jasonNLP/TAT-R1}}.
\end{abstract}

\section{Introduction}

Terminology translation is an essential task in machine translation, and its accuracy significantly impacts the translation quality of specialized domain texts. Many researchers have conducted extensive studies on terminology translation, proposing various methodologies. \citet{term1} detects terms, constructs a terminology database, and provides term information via retrieval-augmented generation (RAG) before model translation. \citet{term2} synthesizes bilingual data containing terms, fine-tunes the model, and applies post-processing to correct terminology after translation. DragFT \citep{dragft} employs few-shot examples to enhance translation performance in specialized domains. \citet{tat} improves term translation by constraining incorrect terminology during decoding. These methods generally rely on relatively accurate terminology extraction to either: 1) construct training data for supervised fine-tuning, or 2) incorporate relevant terminological information during the inference phase.

Recent advances have demonstrated promising progress in leveraging reinforcement learning (RL) to stimulate models' deep reasoning capabilities, exemplified by DeepSeek-R1 \citep{deepseek-r1}. These developments have further validated that the enhanced model abilities acquired through RL exhibit strong generalization performance. Inspired by DeepSeek-R1 \citep{deepseek-r1}, some works have tried to use reinforcement learning to stimulate the model's deep reasoning capabilities and improve translation quality. R1-T1 \citep{R1-T1} synthesizes training data with reasoning processes for translation, first applying SFT and then conducting reinforcement training using COMET\citep{comet22} as the reward. Similar to DeepSeek-R1-Zero, MT-R1-Zero \citep{MT-R1-Zero} directly performs reinforcement training on a pretrained model, employing BLEU\cite{bleu} and COMETKiwi\cite{cometkiwi} as rewards. DeepTrans \citep{Deeptrans} directly uses DeepSeek-V3 \citep{deepseekv3} scoring as the reward, enhancing the model's performance in literary translation through reinforcement learning. To the best of our knowledge, no existing research has explored the integration of reinforcement learning and deep reasoning for terminology translation tasks.

In this paper, we propose TAT-R1, a terminology-aware translation model trained with reinforcement learning and word alignment. First, using word alignment techniques, we design effective reinforcement learning reward signals for terminology translation tasks. Word alignment involves analyzing parallel bilingual corpora to determine translational equivalence between words across languages. By leveraging word alignment techniques, we can effectively extract domain-specific key terms from parallel training corpora, thereby substantially mitigating the challenge of scarce terminology-labeled training data. Then, we directly train our model using RL, and extensive experiment results demonstrate the effectiveness of our proposed method. Our main contributions are as follows:

\begin{itemize}
	\item We propose TAT-R1, the first terminology-aware translation model trained with RL and word alignment rewards. Leveraging word alignment, we design three simple yet effective reward functions for terminology translation model training. 
	\item Experimental results demonstrate the effectiveness of TAT-R1. TAT-R1 significantly improves terminology translation accuracy compared to the baseline while maintaining comparable performance on general translation tasks. Moreover, we do not need any terminology detections during inference.
	\item We conduct detailed ablation studies of the DeepSeek-R1-like training paradigm for machine translation, and reveal several key findings, including the generalization capability of RL, the different impacts of various rewards, and the effectiveness of the reasoning process.
\end{itemize}

\section{Methods}

\begin{figure*}[ht]
	\centering
	\includegraphics[width=1\linewidth]{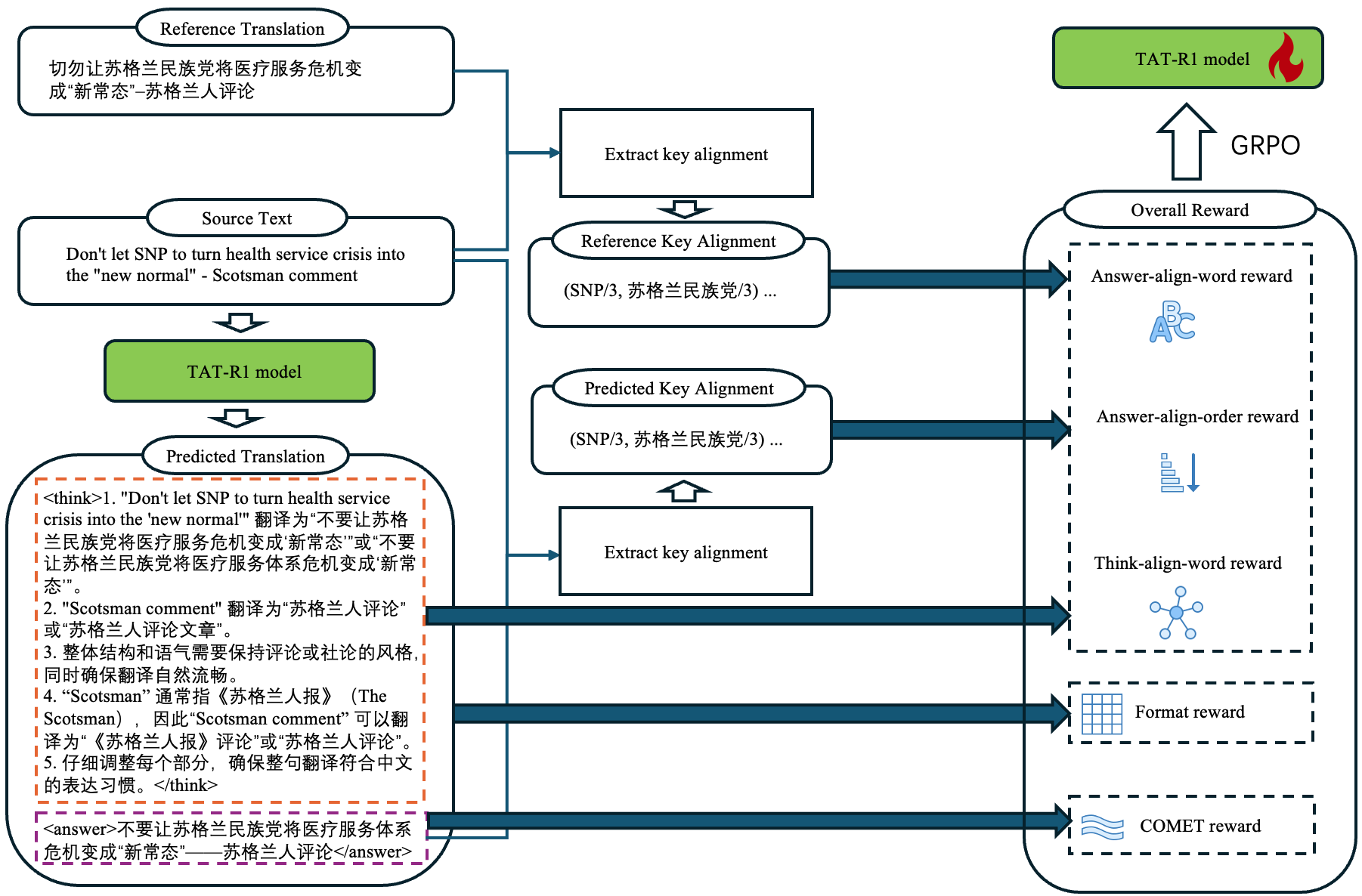}
	\caption{The overview of TAT-R1 training with RL and word alignment.}
	\label{pipline}
\end{figure*}

In this section, we present the reward mechanisms and reinforcement learning algorithm employed in our proposed TAT-R1 model.
\subsection{Rewards With Design}
\label{reward}
As shown in the Figure \ref{pipline}, the rewards we use in our RL training have three parts: format reward, comet reward, and word alignment reward.

\noindent\textbf{Format Reward.}  As shown below, we employed a template similar to that used in DeepSeek-R1, requiring the model to output its reasoning process within <think></think> tags and the translation results within <answer></answer> tags. Here, target\_language specifies the language to translate, while source\_text denotes the input text. To prevent the model from generating non-translation content in the <answer></answer> section, we explicitly included the instruction "without additional explanations" in the User prompt.

\begin{tcolorbox}[
	colframe=gray!80!black, 
	colback=gray!10!white, 
	coltitle=white, 
	fonttitle=\bfseries, 
	title=Template for TAT-R1, 
	boxrule=0.5mm, 
	]

	A conversation between User and Assistant. The User asks a question, and the Assistant solves it. The Assistant first thinks about the reasoning process in the mind and then provides the User with the answer. The reasoning process is enclosed within <think> </think> and answer is enclosed within <answer> </answer> tags, respectively, i.e., <think> reasoning process here </think> <answer> answer here </answer>. \\
	\\
	User: \\
	Translate the following text into \{target\_language\} without additional explanations: \\
	\{source\_text\} \\
	\\
	Assistant:
	\label{template}
\end{tcolorbox}
We employ regular expressions to verify whether the model's output conforms to the format specified in the template. If compliant, the format reward is set to 1; otherwise, it is set to 0. Specifically:

\begin{equation}
R_{format}= \begin{cases}\text{1}, & \text{if the format is right} \\
	\text{0}, & \text{if the format is wrong}\end{cases}
\end{equation}

\noindent\textbf{COMET Reward.} COMET is a widely-used evaluation metric in machine translation that assesses translation quality at the semantic level. The effectiveness of COMET-based rewards has been validated in papers in \citep{R1-T1} and \citep{MT-R1-Zero}. In this work, we incorporate COMET-22 as one component of our reward functions. To maintain training stability, we adopt a similar approach to that used in \citep{R1-T1}, specifically:

\begin{equation}
R_{comet}=round(comet, 2)
\end{equation}

\noindent\textbf{Word Alignment Reward.} Semantic evaluation metrics like COMET primarily assess the overall translation quality of a model but often fail to accurately capture the translation accuracy of localized information, such as technical terms. BLEU, an n-gram-based metric, mainly measures the n-gram overlap between reference translations and model outputs. However, for translation tasks, BLEU imposes overly strict requirements. Unlike mathematics or code, where there is a single correct answer, translation permits multiple valid renditions. Enforcing strict n-gram matching between model outputs and references—as BLEU does—may not always be reasonable and could even introduce negative semantic effects, as we later verify in our experiments. For instance, a single Chinese sentence may have multiple valid English translations with varying syntactic structures.

Nevertheless, key elements like terminology often demand precise translation. By incorporating reward signals that specifically evaluate the accuracy of such critical terms, we can enhance their translation fidelity without compromising overall semantic quality.

In machine translation, word alignment is a critical task that aims to automatically establish correspondences between words in source and target language sentences. This task involves analyzing parallel bilingual corpora to determine translational equivalence between words across languages. In this work, we leverage word alignment to design three distinct reward mechanisms for improving translation quality. Next, we present the detailed computation process for the three word-alignment-based reward mechanisms.

First, we can use word alignment models to identify word-level alignment information between the source text, reference text, and translated text.  

Assume the tokenized sequence of the source text is $S$ and $s_{i}$ is the $i$-th word in $S$. Therefore, the tokenized sequence of the source text can be represented as:
\begin{equation}
	S=[s_1, s_2, s_3, ..., s_i, ... , s_N]
\end{equation}
where $N$ represents the  length of source sequence.
Similarly, the tokenized sequence of the corresponding reference translation and predicted translation can be represented as:
\begin{equation}
	R=[r_1, r_2, r_3, ..., r_j, ..., r_M]
\end{equation}	
\begin{equation}
	P=[p_1, p_2, p_3, ..., p_k, ... , p_K]
\end{equation}
where $M$ represents the  length of reference tokens and $K$ represents the  length of predicted tokens.
Given the tokenized sequences of the source text, reference translation, and predicted translation we can input them into the word alignment model to obtain token-level alignment relationships:
\begin{equation}
	Aref = Align(S, R) = [..., Aref_{ij}, ...]
\end{equation}

\begin{equation}
	Apre = Align(S, P) = [..., Apre_{ik}, ...]
\end{equation}
where $Aref$ represents the alignments between source and reference tokens, and $Apre$ represents the alignments between source and predicted tokens. $Aref_{ij}$ represents the $i$-th token in source tokens and the $j$-th token in reference tokens is aligned. $Apre_{ik}$ represents the $i$-th token in source tokens and the $k$-th token in predicted tokens is aligned, which can be expressed in formulas respectively as $Aref_{ij}=(s_i/i, r_j/j)$ and  $Apre_{ik}=(s_i/i, p_k/k)$.

Next, we perform Named Entity Recognition (NER) on the source tokens, retaining nouns as key elements requiring alignment. This is because noun translations typically exhibit less variability compared to sentence structures or conjunctions, where multiple valid translations often exist. So, after this step, the key alignments are a subset of origin alignments, and the aligned source tokens are all nouns. We denote the set of key alignments as $ Aref_key$ and $ Apre_key$.

Last, we can calculate the alignment reward with $Aref\_key$ and $Apre\_key$. As shown in Figure \ref{pipline}, we design three types of alignment rewards, namely answer-align-word reward $R_{aaw}$,  answer-align-order reward $R_{aao}$, and think-align-word reward $R_{taw}$. 

Answer-align-word reward reflects the word overlap ratio between the reference key alignment and the predicted key alignment. This reward encourages the model to translate key information in its output accurately and can be denoted as:

\begin{equation}
	R_{aaw} = \frac{len(Aref\_key  \cap Apre\_key)}{len(S) + len(P)}
\end{equation}

\noindent\ we include the length of the model's output token sequence in the denominator to prevent the model from generating excessively long outputs to hack this reward.

The answer-align-order reward is a reward that reflects the order overlap ratio between the reference key alignment and the predicted key alignment. This reward encourages the model to follow the order of key information as it appears in the reference translation and can be denoted as:

\begin{equation}
	R_{aao} = \frac{len(OD(Aref\_key) \cap (OD(Apre\_key))}{len(OD(Aref\_key))}
\end{equation}

\noindent where $OD(X)$ means getting the order pairs of sequence $X$. For example, if $X = [a, b, c]$, then $OD(x)=\{ab, ac, bc\}$.

Similar to the answer-align-word reward, the think-align-word reward is a reward that reflects the word overlap ratio between the reference key alignment and the text in tags <think>  and </think>. This reward encourages the model to consider how to translate key information before output the final answer and can be denoted as:

\begin{equation}
	R_{taw} = \frac{num\ of\ Aref\_key\ hit\ in\ think}{len(Aref\_key)}
\end{equation}

\noindent where $num\ of\ Aref\_key\ hit\ in\ think$ means the appeared number of  aligned word pairs in text between <think> and </think> tags. For example, if one  item of $Aref\_key$ is $Aref\_key_{ij}=(s_i/i, r_j/j)$ and both words $s_i$ and $r_j$ appears in think process, then the $num\ of\ Aref\_key\ hit\ in\ think$ should plus one.

\noindent\textbf{Overall Reward.} Given the above rewards, the overall reward we design can be denoted as:

\begin{equation}
	R_{all} = \begin{cases}
		0, & \text{if } R_{format} = 0 \\
		R_{comet} + \alpha*R_{aaw} \\
		\quad + \beta*R_{aao}  \\
		\quad + \gamma*R_{taw}, & \text{if } R_{format} = 1
	\end{cases}
	\label{reward_oevalall}
\end{equation}

\noindent where the hyperparameters $\alpha$, $\beta$ and $\gamma$ control the trade-off between different reward components.

\subsection{RL Algorithom}
Our translation model is trained using the Group Relative Policy Optimization(GRPO) methods \citep{ds-math}, which optimizes policies through a hybrid reward function proposed in Section \ref{reward}.  During training, for each input question $q$, we generate a set of candidate outputs $\{o_1, o_2, ..., o_G\}$ from the current policy model $\pi_{\theta_{old}}$. The advantage value $A_i$ for each candidate is calculated by normalizing its reward $r_i$ against the group’s mean and standard deviation:
\begin{equation}
	A_i = \frac{r_i - \operatorname{mean}(\{r_1, r_2, \dots, r_G\})}{\operatorname{std}(\{r_1, r_2, \dots, r_G\})}
\end{equation}

GRPO then optimizes the policy parameters $\theta$ by maximizing the following objective:

\begin{equation}
	\begin{aligned}
		J_{\mathrm{GRPO}}(\theta) 
		&= \mathbb{E}_{q \sim P(Q),\, \{o_i\}_{i=1}^G \sim \pi_{\theta_{\mathrm{old}}}(O \mid q)} \\
		&\Biggl[
		\frac{1}{G} \sum_{i=1}^G
		\min\!\Bigl(
		\frac{\pi_{\theta}(o_i \mid q)}{\pi_{\theta_{\mathrm{old}}}(o_i \mid q)}\,A_i,\, \\
		&\mathrm{clip}\!\Bigl(
		\frac{\pi_{\theta}(o_i \mid q)}{\pi_{\theta_{\mathrm{old}}}(o_i \mid q)},
		1-\varepsilon,\,
		1+\varepsilon
		\Bigr)
		A_i
		\Bigr)  \\
		&-\,\beta\,D_{\mathrm{KL}}\bigl(\pi_{\theta}\,\big\|\,\pi_{\mathrm{ref}}\bigr)
		\Biggr],
	\end{aligned}
\end{equation}

\noindent where:
\begin{itemize}
	\item $\varepsilon$ controls the clipping range for policy updates, ensuring stable training by limiting drastic changes.
	\item $\beta$ scales the KL divergence penalty $D_{KL}$, which constrains the policy from deviating too far from the reference policy $\pi_{ref}$(typically the initial policy). The KL term is approximated as:  $D_{KL}\bigl(\pi_\theta \,\|\, \pi_{\text{ref}}\bigr)
	= \frac{\pi_{\text{ref}}(o_i \mid q)}{\pi_\theta(o_i \mid q)}
	- \log\!\Bigl(\frac{\pi_{\text{ref}}(o_i \mid q)}{\pi_\theta(o_i \mid q)}\Bigr)
	- 1 \,$
\end{itemize}

\noindent These hyperparameters balance exploration and stability, following prior work in proximal policy optimization \citep{ppo, ds-math}

\section{Experiments and Results}
In this section, we will introduce the relevant experimental setup, present the corresponding experimental results, and provide ablation studies. 
\subsection{Experimetal Setups}
\textbf{Backbone.} We chose Qwen2.5-7B-Instruct \citep{Qwen2.5} as the backbone model because it demonstrates strong multilingual performance among open-source models of comparable parameter size. This helps minimize potential negative impacts caused by insufficient capabilities of the base model.

\noindent\textbf{Datasets.} Following MT-R1-Zero \citep{MT-R1-Zero}, we used Chinese (ZH) to/from English (EN) parallel data from WMT 2017 to WMT 2020 as our training data. Additionally, we incorporated ZH-EN translation pairs from Flores-200 \citep{flores200} and NTREX \citep{ntrex}, resulting in 16,124 training samples.

We selected the ZH-EN test sets from WMT23 and WMT24 for general translation quality evaluation. For terminology-specific translation quality evaluation, we adopted the RTT test set \citep{rtt}, a challenging English->German terminology test set containing 500 sentence pairs.

\noindent\textbf{Evaluation Metrics.} For general translation quality evaluation, we choose BLEU \citep{bleu,bleu2}, COMETKiwi-23-XL \citep{cometkiwi}, and XCOMET-XL \citep{xcomet}. BLEU is a lexical metric. COMETKiwi is a reference-free learning-based metric. XCOMET is a reference-based learning metric. These three metrics complement each other to some extent, enabling the evaluation of translation quality at both the lexical and semantic levels.

For terminology translation quality evaluation, in addition to the three metrics mentioned above, we also assess terminology accuracy (TA), indicating how many of the source terms have a corresponding target term in the translation.

\noindent\textbf{Word Alignment.} We select the open-source model SimAlign \citep{simalign} to extract word alignment information between the source text and the translation. SimAlign is an unsupervised word alignment tool with strong performance in word alignment tasks across multiple language pairs, including English and Chinese. SimAlign takes the tokenized sequences of the texts to be aligned as input. For Chinese, we use Jieba\footnote{\url{https://github.com/fxsjy/jieba}} for word segmentation, while for English, we employ NLTK\footnote{\url{https://www.nltk.org/}}.

\noindent\textbf{Training Details.} We conduct our training based on the verl \footnote{\url{https://github.com/volcengine/verl}} framework. For the hyperparameters in overall reward, we set $\alpha$, $\beta$, and $\gamma$ to $1$, $\frac{1}{10}$, and $\frac{1}{10}$, respectively. In the GRPO algorithm, we set the number of rollouts to 16, the sampling temperature to 1.0, and use a constant learning rate of 1e-6. The maximum generation length is 4,096 tokens, with a training batch size of 128. All experiments are trained for three epochs.

\begin{table*}[]
	\centering
	\scriptsize
	\renewcommand\arraystretch{1.2}
	\begin{tabular}{p{4cm}|cccc|cccc}
		\hline
		\multirow{2}{*}{\textbf{Models}} & \multicolumn{4}{c|}{\textbf{ZH-\textgreater{}EN}} & \multicolumn{4}{c}{\textbf{EN-\textgreater{}ZH}} \\ \cline{2-9} 
		& \multicolumn{1}{c|}{\textbf{BLEU}} & \multicolumn{1}{c|}{\textbf{COMETKiwi}} & \multicolumn{1}{c|}{\textbf{XCOMET}} & \textbf{Avg.} & \multicolumn{1}{c|}{\textbf{BLEU}} & \multicolumn{1}{c|}{\textbf{COMETKiwi}} & \multicolumn{1}{c|}{\textbf{XCOMET}} & \textbf{Avg.} \\ \hline
		Qwen2.5-7B-Instruct & \multicolumn{1}{c|}{22.18} & \multicolumn{1}{c|}{73.08} & \multicolumn{1}{c|}{86.05} & 60.44 & \multicolumn{1}{c|}{37.36} & \multicolumn{1}{c|}{71.65} & \multicolumn{1}{c|}{75.46} & 61.49 \\ \hline
		SFT & \multicolumn{1}{c|}{22.15} & \multicolumn{1}{c|}{73.98} & \multicolumn{1}{c|}{86.27} & 60.80 & \multicolumn{1}{c|}{33.42} & \multicolumn{1}{c|}{68.58} & \multicolumn{1}{c|}{75.19} & 59.06 \\ \hline
		RL-$R_{comet}$ & \multicolumn{1}{c|}{22.32} & \multicolumn{1}{c|}{77.51} & \multicolumn{1}{c|}{88.59} & 62.80 & \multicolumn{1}{c|}{36.12} & \multicolumn{1}{c|}{75.79} & \multicolumn{1}{c|}{79.42} & 63.78 \\ \hline
		RL-$R_{comet}+R_{BLEU}$ & \multicolumn{1}{c|}{25.08} & \multicolumn{1}{c|}{75.83} & \multicolumn{1}{c|}{87.62} & 62.84 & \multicolumn{1}{c|}{40.98} & \multicolumn{1}{c|}{71.33} & \multicolumn{1}{c|}{77.08} & 63.13 \\ \hline
		RL-$R_{comet}+R_{aaw}$ & \multicolumn{1}{c|}{23.90} & \multicolumn{1}{c|}{77.39} & \multicolumn{1}{c|}{88.37} & 63.22 & \multicolumn{1}{c|}{39.05} & \multicolumn{1}{c|}{73.52} & \multicolumn{1}{c|}{78.51} & 63.69 \\ \hline
		RL-$R_{comet}+R_{aaw}+R_{aao}$ & \multicolumn{1}{c|}{23.97} & \multicolumn{1}{c|}{77.21} & \multicolumn{1}{c|}{88.27} & 63.15 & \multicolumn{1}{c|}{38.53} & \multicolumn{1}{c|}{74.94} & \multicolumn{1}{c|}{78.65} & 64.04 \\ \hline
		RL-$R_{all}$ (TAT-R1) & \multicolumn{1}{c|}{24.40} & \multicolumn{1}{c|}{77.20} & \multicolumn{1}{c|}{88.38} & \textbf{63.33} & \multicolumn{1}{c|}{39.45} & \multicolumn{1}{c|}{75.57} & \multicolumn{1}{c|}{78.65} & \textbf{64.56} \\ \hline
	\end{tabular}
	\caption{Performance on WMT23 ZH to EN and WMT24 EN to ZH testset. ZH represents Chinese and EN represents English. \textit{Avg.} represents the average of BLEU, COMETKiwi and XCOMET metrics.}
	\label{performance_of_wmt}
\end{table*}

\begin{table}[]
	\centering
	\scriptsize
	\renewcommand\tabcolsep{0.8pt}
	\renewcommand\arraystretch{1.2}
	\begin{tabular}{l|ccccc}
		\hline
		\multirow{2}{*}{\textbf{Models}} & \multicolumn{5}{c}{\textbf{EN-\textgreater{}DE}} \\ \cline{2-6} 
		& \multicolumn{1}{c|}{\textbf{BLEU}} & \multicolumn{1}{c|}{\textbf{COMETKiwi}} & \multicolumn{1}{c|}{\textbf{XCOMET}} & \multicolumn{1}{c|}{\textbf{TA}} & \textbf{Avg.} \\ \hline
		Qwen2.5-7B-Instruct & \multicolumn{1}{c|}{25.87} & \multicolumn{1}{c|}{67.05} & \multicolumn{1}{c|}{88.65} & \multicolumn{1}{c|}{53.29} & 58.72 \\ \hline
		SFT & \multicolumn{1}{c|}{0.00} & \multicolumn{1}{c|}{-} &\multicolumn{1}{c|}{-} & \multicolumn{1}{c|}{0.08} & - \\ \hline
		RL-$R_{comet}$ & \multicolumn{1}{c|}{24.52} & \multicolumn{1}{c|}{70.26} & \multicolumn{1}{c|}{90.17} & \multicolumn{1}{c|}{54.42} & 59.84 \\ \hline
		RL-$R_{comet}$+$R_{BLEU}$ & \multicolumn{1}{c|}{27.37} & \multicolumn{1}{c|}{66.49} & \multicolumn{1}{c|}{88.77} & \multicolumn{1}{c|}{54.91} & 59.39 \\ \hline
		RL-$R_{comet}$+$R_{aaw}$ & \multicolumn{1}{c|}{26.21} & \multicolumn{1}{c|}{71.33} & \multicolumn{1}{c|}{90.56} & \multicolumn{1}{c|}{55.57} & 60.92 \\ \hline
		RL-$R_{comet}$+$R_{aaw}$+$R_{aao}$ & \multicolumn{1}{c|}{26.34} & \multicolumn{1}{c|}{72.04} & \multicolumn{1}{c|}{90.99} & \multicolumn{1}{c|}{55.73} & 61.28 \\ \hline
		\begin{tabular}[c]{@{}c@{}}RL-$R_{all}$ (TAT-R1) \end{tabular} & \multicolumn{1}{c|}{27.10} & \multicolumn{1}{c|}{\textbf{73.82}} & \multicolumn{1}{c|}{\textbf{91.22}} & \multicolumn{1}{c|}{\textbf{56.42}} & \textbf{62.14} \\ \hline
	\end{tabular}
	\caption{Performance on RTT testset. DE represents the German language. \textit{Avg.} represents the average of BLEU, COMETKiwi, XCOMET and TA metrics.}
	\label{performance_of_rtt}
\end{table}

\subsection{Results and Analysis}
This section presents the main experimental results, demonstrating that our proposed word-alignment reward is highly effective. We then provide a detailed analysis of the experimental outcomes and supplement the findings with relevant ablation studies.

\subsubsection{Main Results}

Table \ref{performance_of_wmt} presents the performance of models trained under different settings on the WMT test set, reflecting their general Chinese-English translation capabilities. Table \ref{performance_of_rtt} shows the results on the RTT test set, which demonstrate the models' terminology translation abilities. Here, \textit{SFT} denotes the model obtained by fine-tuning Qwen2.5-7B-Instruct with our training data, while \textit{RL-x} represents the model trained through reinforcement learning on Qwen2.5-7B-Instruct using our data, where \textit{x} indicates different rewards employed during the reinforcement process. For example, RL-$R_{comet}+R_{BLEU}$ refers to the model reinforced using both COMET and BLEU as rewards. "Avg." in the table represents the average value of all metrics.

Regarding general translation performance (Table \ref{performance_of_wmt}), our model TAT-R1 significantly improves across all metrics compared to the baseline Qwen2.5-7B-Instruct. For ZH→EN translation, the average metric increased from 60.44 to 63.33 (a 2.99\% improvement), while for EN→ZH, it rises from 61.49 to 64.56 (a 3.07\% improvement). Compared to using only COMET as a reward (RL-$R_{comet}$), incorporating three word alignment-related rewards further enhanced the model's overall performance metrics.

As shown in Table \ref{performance_of_rtt}, on the terminology test set RTT, our model TAT-R1 with word alignment rewards demonstrates significant improvements over RL-$R_{comet}$ (without word alignment rewards) across all evaluation metrics: BLEU score increased by 2.58\%, COMETKiwi by 3.56\%, XCOMET by 1.05\%, and terminology accuracy(TA) by 2\%.

Compared to models not reinforced with word alignment information, our TAT-R1 achieves slightly better performance in general translation tasks and significantly superior results in terminology translation, demonstrating the effectiveness of our proposed word alignment reward mechanism.

\subsubsection{Ablation Study}

\begin{figure}[h]
	\centering
	\includegraphics[width=1\columnwidth]{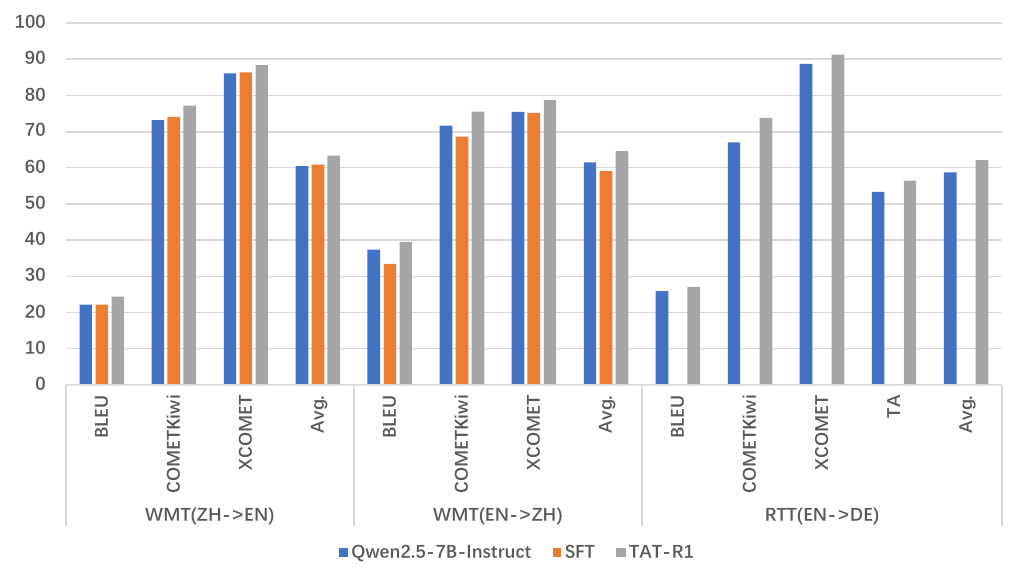}
	\caption{Compare the performance between SFT and RL.}
	\label{compare_sft_rl}
\end{figure}

\begin{figure*}[ht]
	\centering
	\includegraphics[width=1\linewidth]{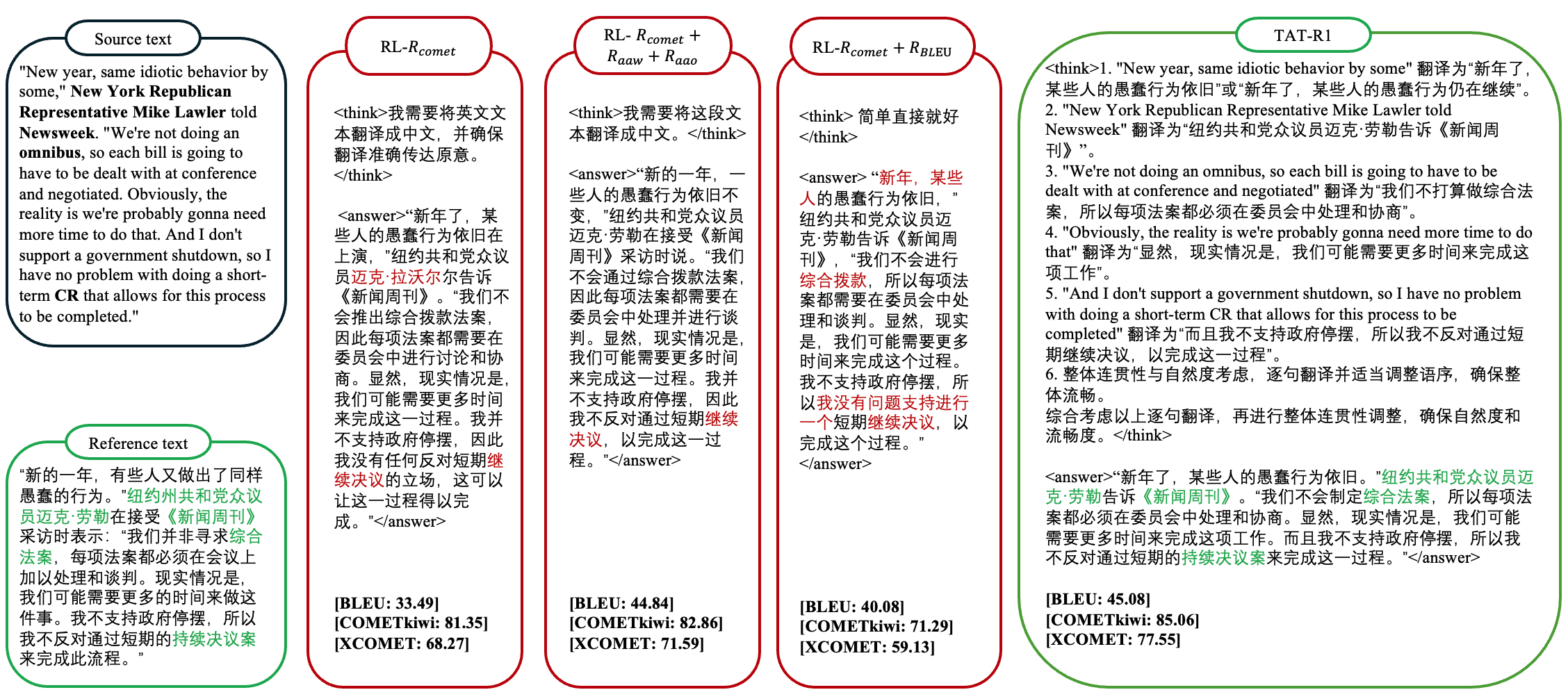}
	\caption{Qualitative examples illustrate the effect of different rewards on EN to ZH translation.}
	\label{case_studies}
\end{figure*}

\begin{figure}[h]
	\centering
	\includegraphics[width=1\columnwidth]{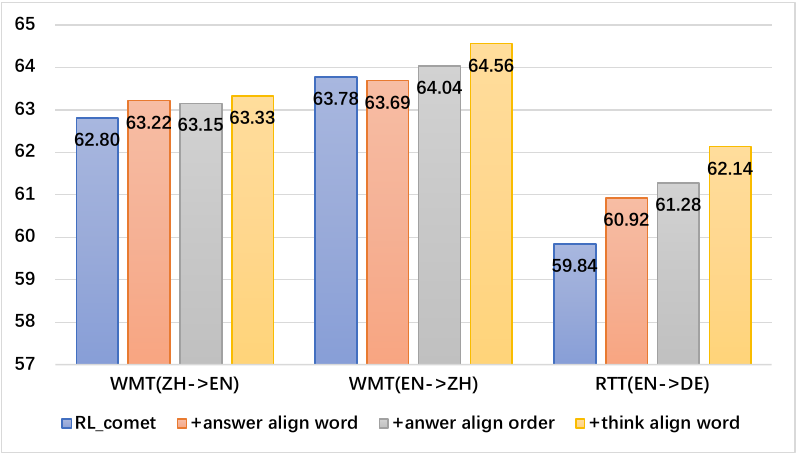}
	\caption{Compare the average performance between different word alignment rewards.}
	\label{compare_align_reward}
\end{figure}

\noindent\textbf{Compare between SFT and RL.} To demonstrate the effectiveness of RL, we fine-tune the model using the same training data with SFT. As shown in Figure \ref{compare_sft_rl}, although the fine-tuned model showed a slight improvement over the baseline in Chinese-to-English (Zh->En) translation on WMT, there was a noticeable decline in English-to-Chinese (En->Zh) metrics. We attribute this to noise in the current training data and that not all reference translations are of higher quality than the model's origin outputs, negatively impacting the translation performance after SFT. On the terminology test set RTT, the fine-tuned model almost entirely mistranslates English into Chinese for the English-to-German (En->De) task, resulting in all metrics dropping close to zero. In contrast, the RL-trained TAT-R1 model improved across all metrics, demonstrating strong performance on the out-of-distribution (OOD) En->De task. This phenomenon indicates that, in translation tasks, RL-trained models exhibit better stability and generalization capabilities compared to SFT.

\begin{table}[]
	\centering
	\scriptsize
	\renewcommand\arraystretch{1.2}
	\begin{tabular}{|c|c|c|c|c|}
		\hline
		\textbf{Datasets} & \textbf{Metrics} & \textbf{\makecell{RL\\comet}} & \textbf{\makecell{RL\\comet+bleu}} & \textbf{TAT-R1} \\ \hline
		\multirow{4}{*}{\textbf{\makecell{WMT\\(ZH-\textgreater{}EN)}}} & \textbf{BLEU} & 22.32 & 25.08 & 24.40 \\ \cline{2-5} 
		& \textbf{COMETKiwi} & 77.51 & 75.83 & 77.20 \\ \cline{2-5} 
		& \textbf{XCOMET} & 88.59 & 87.62 & 88.38 \\ \cline{2-5} 
		& \textbf{Avg.} & 62.80 & 62.84 & 63.33 \\ \hline
		\multirow{4}{*}{\textbf{\makecell{WMT\\(EN-\textgreater{}ZH)}}} & \textbf{BLEU} & 36.12 & 40.98 & 39.45 \\ \cline{2-5} 
		& \textbf{COMETKiwi} & 75.79 & 71.33 & 75.57 \\ \cline{2-5} 
		& \textbf{XCOMET} & 79.42 & 77.08 & 78.65 \\ \cline{2-5} 
		& \textbf{Avg.} & 63.78 & 63.13 & 64.56 \\ \hline
		\multirow{5}{*}{\textbf{\makecell{RTT\\(EN-\textgreater{}DE)}}} & \textbf{BLEU} & 24.52 & 27.37 & 27.10 \\ \cline{2-5} 
		& \textbf{COMETKiwi} & 70.26 & 66.49 & 73.82 \\ \cline{2-5} 
		& \textbf{XCOMET} & 90.17 & 88.77 & 91.22 \\ \cline{2-5} 
		& \textbf{TA} & 54.42 & 54.91 & 56.42 \\ \cline{2-5} 
		& \textbf{Avg.} & 59.84 & 59.39 & 62.14 \\ \hline
	\end{tabular}
	\caption{Compare between BLEU and word alignment rewards.}
	\label{compare_bleu}
\end{table}

\noindent\textbf{The Effects of Different Word Alignment Rewards.} Figure \ref{compare_align_reward} demonstrates the average performance of the model when incrementally incorporating answer-align-word reward, answer-align-order reward, and think-align-word reward based on RL-$R_{comet}$. The results show that with the addition of each word alignment reward, the model's performance consistently improves, validating the effectiveness of our proposed word alignment rewards.

In Figure \ref{case_studies}, we also present some output cases of the model after applying different rewards. We observe that when rewards are calculated only for the model's output within <answer></answer>, the final "think" step often produces non-functional statements like "I need to translate the English text into Chinese and ensure the translation accurately conveys the original meaning."—failing to generate meaningful reasoning. This is evident in the RL-$R_{comet}$, RL-$R_{comet}+R_{aaw}$, and RL-$R_{comet}+R_{BLEU}$ examples in Figure \ref{case_studies}. However, after introducing the think word alignment reward, the model begins to reason about the translation of key information in the "think" step, leading to a significant improvement in the final metrics, as shown in the TAT-R1 example in Figure \ref{case_studies}.

\noindent\textbf{Compare between BLEU and Word Alignment Rewards.} As shown in Table \ref{compare_bleu}, while the BLEU reward significantly improves the BLEU metric, it has a notably negative impact on semantic evaluation metrics such as COMET. We further analyze specific cases (e.g., RL-$R_{comet}+R_{BLEU}$ in Figure \ref{case_studies}) and find that models trained with BLEU as the reward exhibit an apparent degradation in translation fluency. In contrast, the word alignment rewards focus solely on the correctness of keyword translations, demonstrating positive effects on lexical and semantic translation quality.

\section{Related Work}

\subsection{Reason-based LLMs}
In recent years, reason-based large language models, such as OpenAI's o1  \citep{openai-o1} and DeeeSeek-R1(\citep{deepseek-r1}), have demonstrated strong performance across various tasks, attracting significant attention from researchers. Recent studies primarily focus on solving complex reasoning tasks, such as mathematical problem-solving and code generation \cite{zeng2025simplerl, OpenReasonerZero2025, deepscaler2025, fastcurl, o1replicationjourney, o1-coder}. However, recent efforts have increasingly explored applying reason-based LLMs to general tasks. For instance, marco-o1 \citep{marco-o1} investigates the use of reasoning-enhanced models in open-ended text generation, where there are no clear-cut standards for evaluating correctness, unlike in mathematics or programming. Some surveys \citep{surveyTowardsReasoningEra, surveyFromSystem1toSystem2} provide systematic reviews of the advancements and trends in reason-based LLMs. 

\subsection{Reason-based LLMs for MT}
Some researchers have attempted to explore the capabilities of reason-based LLMs in machine translation tasks. Marco-o1 \citep{marco-o1} and \citet{Newtrends} briefly demonstrate that reason-based LLMs can somewhat improve translation performance. DRT \citep{DRT} enhances the model's effectiveness in literary translation by synthesizing translation data with reasoning processes and performing supervised fine-tuning (SFT). \citet{Evaluatingo1-LikeLLMs} provides a preliminary assessment of the performance of multiple reason-based LLMs in machine translation. Inspired by DeepSeek-R1 \citep{deepseek-r1}, some works have tried to use reinforcement learning to stimulate the model's deep reasoning capabilities and improve translation quality. R1-T1 \citep{R1-T1} synthesizes training data with reasoning processes for translation, first applying SFT and then conducting reinforcement training using COMET as the reward. Like DeepSeek-R1-Zero, MT-R1-Zero \citep{MT-R1-Zero} directly performs reinforcement training on a pretrained model, employing BLEU and COMET as rewards. DeepTrans \citep{Deeptrans} directly uses DeepSeek-V3 scoring as the reward, enhancing the model's performance in literary translation through reinforcement learning.

\subsection{Terminology Translation}
In many fields, accurate translation of terminology is crucial. In recent years, numerous researchers have explored terminology translation using LLMs. \citet{term1} detects terms, constructs a terminology database, and provides term information via retrieval-augmented generation (RAG) before model translation. \citet{term2} synthesizes bilingual data containing terms, fine-tunes the model, and applies post-processing to correct terminology after translation. For technical terms, \citet{term3} proposes a parenthetical terminology translation method. DragFT \citep{dragft} employs few-shot examples to enhance translation performance in specialized domains. \citet{tat} improves term translation by constraining incorrect terminology during decoding. To better evaluate models' terminology translation capabilities, \citet{rtt} introduces a new terminology test set and examines the effects of various data augmentation methods on term translation.

\section{Conclusion}

In this work, we introduce TAT-R1, the first terminology-aware translation model trained with RL and word alignment. Empowered by word alignment in machine translation, we design three types of new rule-based rewards. Combining the word alignment rewards with format reward and comet reward, we train our model with GRPO. Experimental results demonstrate the effectiveness of TAT-R1. TAT-R1 significantly improves terminology translation accuracy compared to the baseline while maintaining comparable performance on general translation tasks.

\section*{Limitations}

While TAT-R1 has significantly improved terminology translation accuracy, certain limitations remain. The reasoning process we observe is relatively simple, and we have not observed complex reasoning processes, such as self-correction and verification, which appear in mathematical tasks. This discrepancy may reflect the differences between the machine translation task and the mathematical task or indicate the need for specialized design in machine translation tasks. Another limitation is that we have not systematically explored multiple translation evaluation metrics as potential rewards, such as BLEURT \citep{bleurt}, MetricX \citep{metricx}, and GEMBA \citep{gemba}. A promising future research direction would be investigating diverse reward signals for translation quality assessment, combined with word-alignment-based rewards, to validate their effectiveness in terminology translation tasks further.

\bibliography{custom}

\begin{thebibliography}{38}
\providecommand{\natexlab}[1]{#1}

\bibitem[{Bogoychev and Chen(2023)}]{tat}
Nikolay Bogoychev and Pinzhen Chen. 2023.
\newblock \href {https://doi.org/10.18653/V1/2023.WMT-1.80} {Terminology-aware
  translation with constrained decoding and large language model prompting}.
\newblock In \emph{Proceedings of the Eighth Conference on Machine Translation,
  {WMT} 2023, Singapore, December 6-7, 2023}, pages 890--896. Association for
  Computational Linguistics.

\bibitem[{Chen et~al.(2025{\natexlab{a}})Chen, Song, Zhu, Chen, Yang, Zhao, and
  Zhang}]{Evaluatingo1-LikeLLMs}
Andong Chen, Yuchen Song, Wenxin Zhu, Kehai Chen, Muyun Yang, Tiejun Zhao, and
  Min Zhang. 2025{\natexlab{a}}.
\newblock \href {https://doi.org/10.48550/ARXIV.2502.11544} {Evaluating o1-like
  llms: Unlocking reasoning for translation through comprehensive analysis}.
\newblock \emph{CoRR}, abs/2502.11544.

\bibitem[{Chen et~al.(2025{\natexlab{b}})Chen, Qin, Liu, Peng, Guan, Wang, Hu,
  Zhou, Gao, and Che}]{surveyTowardsReasoningEra}
Qiguang Chen, Libo Qin, Jinhao Liu, Dengyun Peng, Jiannan Guan, Peng Wang,
  Mengkang Hu, Yuhang Zhou, Te~Gao, and Wanxiang Che. 2025{\natexlab{b}}.
\newblock \href {https://doi.org/10.48550/ARXIV.2503.09567} {Towards reasoning
  era: {A} survey of long chain-of-thought for reasoning large language
  models}.
\newblock \emph{CoRR}, abs/2503.09567.

\bibitem[{Costa{-}juss{\`{a}} et~al.(2022)Costa{-}juss{\`{a}}, Cross,
  {\c{C}}elebi, Elbayad, Heafield, Heffernan, Kalbassi, Lam, Licht, Maillard,
  Sun, Wang, Wenzek, Youngblood, Akula, Barrault, Gonzalez, Hansanti, Hoffman,
  Jarrett, Sadagopan, Rowe, Spruit, Tran, Andrews, Ayan, Bhosale, Edunov, Fan,
  Gao, Goswami, Guzm{\'{a}}n, Koehn, Mourachko, Ropers, Saleem, Schwenk, and
  Wang}]{flores200}
Marta~R. Costa{-}juss{\`{a}}, James Cross, Onur {\c{C}}elebi, Maha Elbayad,
  Kenneth Heafield, Kevin Heffernan, Elahe Kalbassi, Janice Lam, Daniel Licht,
  Jean Maillard, Anna~Y. Sun, Skyler Wang, Guillaume Wenzek, Al~Youngblood,
  Bapi Akula, Lo{\"{\i}}c Barrault, Gabriel~Mejia Gonzalez, Prangthip Hansanti,
  John Hoffman, and 19 others. 2022.
\newblock \href {https://doi.org/10.48550/ARXIV.2207.04672} {No language left
  behind: Scaling human-centered machine translation}.
\newblock \emph{CoRR}, abs/2207.04672.

\bibitem[{DeepSeek{-}AI et~al.(2025)DeepSeek{-}AI, Guo, Yang, Zhang, Song,
  Zhang, Xu, Zhu, Ma, Wang, Bi, Zhang, Yu, Wu, Wu, Gou, Shao, Li, Gao, Liu,
  Xue, Wang, Wu, Feng, Lu, Zhao, Deng, Zhang, Ruan, Dai, Chen, Ji, Li, Lin,
  Dai, Luo, Hao, Chen, Li, Zhang, Bao, Xu, Wang, Ding, Xin, Gao, Qu, Li, Guo,
  Li, Wang, Chen, Yuan, Qiu, Li, Cai, Ni, Liang, Chen, Dong, Hu, Gao, Guan,
  Huang, Yu, Wang, Zhang, Zhao, Wang, Zhang, Xu, Xia, Zhang, Zhang, Tang, Li,
  Wang, Li, Tian, Huang, Zhang, Wang, Chen, Du, Ge, Zhang, Pan, Wang, Chen,
  Jin, Chen, Lu, Zhou, Chen, Ye, Wang, Yu, Zhou, Pan, and Li}]{deepseek-r1}
DeepSeek{-}AI, Daya Guo, Dejian Yang, Haowei Zhang, Junxiao Song, Ruoyu Zhang,
  Runxin Xu, Qihao Zhu, Shirong Ma, Peiyi Wang, Xiao Bi, Xiaokang Zhang,
  Xingkai Yu, Yu~Wu, Z.~F. Wu, Zhibin Gou, Zhihong Shao, Zhuoshu Li, Ziyi Gao,
  and 81 others. 2025.
\newblock \href {https://doi.org/10.48550/ARXIV.2501.12948} {Deepseek-r1:
  Incentivizing reasoning capability in llms via reinforcement learning}.
\newblock \emph{CoRR}, abs/2501.12948.

\bibitem[{DeepSeek-AI et~al.(2025)DeepSeek-AI, Liu, Feng, Xue, Wang, Wu, Lu,
  Zhao, Deng, Zhang, Ruan, Dai, Guo, Yang, Chen, Ji, Li, Lin, Dai, Luo, Hao,
  Chen, Li, Zhang, Bao, Xu, Wang, Zhang, Ding, Xin, Gao, Li, Qu, Cai, Liang,
  Guo, Ni, Li, Wang, Chen, Chen, Yuan, Qiu, Li, Song, Dong, Hu, Gao, Guan,
  Huang, Yu, Wang, Zhang, Xu, Xia, Zhao, Wang, Zhang, Li, Wang, Zhang, Zhang,
  Tang, Li, Tian, Huang, Wang, Zhang, Wang, Zhu, Chen, Du, Chen, Jin, Ge,
  Zhang, Pan, Wang, Xu, Zhang, Chen, Li, Lu, Zhou, Chen, Wu, Ye, Ye, Ma, Wang,
  Zhou, Yu, Zhou, Pan, Wang, Yun, Pei, Sun, Xiao, Zeng, Zhao, An, Liu, Liang,
  Gao, Yu, Zhang, Li, Jin, Wang, Bi, Liu, Wang, Shen, Chen, Zhang, Chen, Nie,
  Sun, Wang, Cheng, Liu, Xie, Liu, Yu, Song, Shan, Zhou, Yang, Li, Su, Lin, Li,
  Wang, Wei, Zhu, Zhang, Xu, Xu, Huang, Li, Zhao, Sun, Li, Wang, Yu, Zheng,
  Zhang, Shi, Xiong, He, Tang, Piao, Wang, Tan, Ma, Liu, Guo, Wu, Ou, Zhu,
  Wang, Gong, Zou, He, Zha, Xiong, Ma, Yan, Luo, You, Liu, Zhou, Wu, Ren, Ren,
  Sha, Fu, Xu, Huang, Zhang, Xie, Zhang, Hao, Gou, Ma, Yan, Shao, Xu, Wu,
  Zhang, Li, Gu, Zhu, Liu, Li, Xie, Song, Gao, and Pan}]{deepseekv3}
DeepSeek-AI, Aixin Liu, Bei Feng, Bing Xue, Bingxuan Wang, Bochao Wu, Chengda
  Lu, Chenggang Zhao, Chengqi Deng, Chenyu Zhang, Chong Ruan, Damai Dai, Daya
  Guo, Dejian Yang, Deli Chen, Dongjie Ji, Erhang Li, Fangyun Lin, Fucong Dai,
  and 181 others. 2025.
\newblock \href {https://arxiv.org/abs/2412.19437} {Deepseek-v3 technical
  report}.
\newblock \emph{Preprint}, arXiv:2412.19437.

\bibitem[{Federmann et~al.(2022)Federmann, Kocmi, and Xin}]{ntrex}
Christian Federmann, Tom Kocmi, and Ying Xin. 2022.
\newblock \href {https://doi.org/10.18653/v1/2022.sumeval-1.4} {{NTREX}-128
  {--} news test references for {MT} evaluation of 128 languages}.
\newblock In \emph{Proceedings of the First Workshop on Scaling Up Multilingual
  Evaluation}, pages 21--24, Online. Association for Computational Linguistics.

\bibitem[{Feng et~al.(2025)Feng, Cao, Ren, Su, Chen, Zhang, Xu, Hu, Wu, and
  Liu}]{MT-R1-Zero}
Zhaopeng Feng, Shaosheng Cao, Jiahan Ren, Jiayuan Su, Ruizhe Chen, Yan Zhang,
  Zhe Xu, Yao Hu, Jian Wu, and Zuozhu Liu. 2025.
\newblock \href {https://arxiv.org/abs/2504.10160} {Mt-r1-zero: Advancing
  llm-based machine translation via r1-zero-like reinforcement learning}.
\newblock \emph{Preprint}, arXiv:2504.10160.

\bibitem[{Guerreiro et~al.(2024)Guerreiro, Rei, van Stigt, Coheur, Colombo, and
  Martins}]{xcomet}
Nuno~Miguel Guerreiro, Ricardo Rei, Daan van Stigt, Lu{\'{\i}}sa Coheur, Pierre
  Colombo, and Andr{\'{e}} F.~T. Martins. 2024.
\newblock \href {https://doi.org/10.1162/TACL\_A\_00683} {xcomet : Transparent
  machine translation evaluation through fine-grained error detection}.
\newblock \emph{Trans. Assoc. Comput. Linguistics}, 12:979--995.

\bibitem[{He et~al.(2025)He, Liu, Tao, Luo, Zeng, Su, Zhang, Ma, Wei, Meng,
  Yang, Chen, and Yoshie}]{R1-T1}
Minggui He, Yilun Liu, Shimin Tao, Yuanchang Luo, Hongyong Zeng, Chang Su,
  Li~Zhang, Hongxia Ma, Daimeng Wei, Weibin Meng, Hao Yang, Boxing Chen, and
  Osamu Yoshie. 2025.
\newblock \href {https://doi.org/10.48550/ARXIV.2502.19735} {{R1-T1:} fully
  incentivizing translation capability in llms via reasoning learning}.
\newblock \emph{CoRR}, abs/2502.19735.

\bibitem[{Hu et~al.(2025)Hu, Zhang, Han, Jiang, and
  Xiangyu~Zhang}]{OpenReasonerZero2025}
Jingcheng Hu, Yinmin Zhang, Qi~Han, Daxin Jiang, and Heung-Yeung~Shum
  Xiangyu~Zhang. 2025.
\newblock Open-reasoner-zero: An open source approach to scaling reinforcement
  learning on the base model.
\newblock \url{https://github.com/Open-Reasoner-Zero/Open-Reasoner-Zero}.

\bibitem[{Jaech et~al.(2024)Jaech, Kalai, Lerer, Richardson, El{-}Kishky, Low,
  Helyar, Madry, Beutel, Carney, Iftimie, Karpenko, Passos, Neitz, Prokofiev,
  Wei, Tam, Bennett, Kumar, Saraiva, Vallone, Duberstein, Kondrich, Mishchenko,
  Applebaum, Jiang, Nair, Zoph, Ghorbani, Rossen, Sokolowsky, Barak, McGrew,
  Minaiev, Hao, Baker, Houghton, McKinzie, Eastman, Lugaresi, Bassin, Hudson,
  Li, de~Bourcy, Voss, Shen, Zhang, Koch, Orsinger, Hesse, Fischer, Chan,
  Roberts, Kappler, Levy, Selsam, Dohan, Farhi, Mely, Robinson, Tsipras, Li,
  Oprica, Freeman, Zhang, Wong, Proehl, Cheung, Mitchell, Wallace, Ritter,
  Mays, Wang, Such, Raso, Leoni, Tsimpourlas, Song, von Lohmann, Sulit, Salmon,
  Parascandolo, Chabot, Zhao, Brockman, Leclerc, Salman, Bao, Sheng, Andrin,
  Bagherinezhad, Ren, Lightman, Chung, Kivlichan, O'Connell, Osband, Gilaberte,
  and Akkaya}]{openai-o1}
Aaron Jaech, Adam Kalai, Adam Lerer, Adam Richardson, Ahmed El{-}Kishky, Aiden
  Low, Alec Helyar, Aleksander Madry, Alex Beutel, Alex Carney, Alex Iftimie,
  Alex Karpenko, Alex~Tachard Passos, Alexander Neitz, Alexander Prokofiev,
  Alexander Wei, Allison Tam, Ally Bennett, Ananya Kumar, and 80 others. 2024.
\newblock \href {https://doi.org/10.48550/ARXIV.2412.16720} {Openai o1 system
  card}.
\newblock \emph{CoRR}, abs/2412.16720.

\bibitem[{Juraska et~al.(2024)Juraska, Deutsch, Finkelstein, and
  Freitag}]{metricx}
Juraj Juraska, Daniel Deutsch, Mara Finkelstein, and Markus Freitag. 2024.
\newblock \href {https://doi.org/10.18653/v1/2024.wmt-1.35} {{M}etric{X}-24:
  The {G}oogle submission to the {WMT} 2024 metrics shared task}.
\newblock In \emph{Proceedings of the Ninth Conference on Machine Translation},
  pages 492--504, Miami, Florida, USA. Association for Computational
  Linguistics.

\bibitem[{Kim et~al.(2024)Kim, Sung, Lee, Lim, and Perez}]{term1}
Sejoon Kim, Mingi Sung, Jeonghwan Lee, Hyunkuk Lim, and Jorge~Gimenez Perez.
  2024.
\newblock \href {https://aclanthology.org/2024.wmt-1.51} {Efficient terminology
  integration for llm-based translation in specialized domains}.
\newblock In \emph{Proceedings of the Ninth Conference on Machine Translation,
  {WMT} 2024, Miami, FL, USA, November 15-16, 2024}, pages 636--642.
  Association for Computational Linguistics.

\bibitem[{Kocmi and Federmann(2023)}]{gemba}
Tom Kocmi and Christian Federmann. 2023.
\newblock \href {https://aclanthology.org/2023.eamt-1.19} {Large language
  models are state-of-the-art evaluators of translation quality}.
\newblock In \emph{Proceedings of the 24th Annual Conference of the European
  Association for Machine Translation, {EAMT} 2023, Tampere, Finland, 12-15
  June 2023}, pages 193--203. European Association for Machine Translation.

\bibitem[{Li et~al.(2025)Li, Zhang, Zhang, Zhang, Liu, Yao, Xu, Zheng, Wang,
  Chen, Zhang, Yin, Dong, Guo, Song, and Liu}]{surveyFromSystem1toSystem2}
Zhong{-}Zhi Li, Duzhen Zhang, Ming{-}Liang Zhang, Jiaxin Zhang, Zengyan Liu,
  Yuxuan Yao, Haotian Xu, Junhao Zheng, Pei{-}Jie Wang, Xiuyi Chen, Yingying
  Zhang, Fei Yin, Jiahua Dong, Zhijiang Guo, Le~Song, and Cheng{-}Lin Liu.
  2025.
\newblock \href {https://doi.org/10.48550/ARXIV.2502.17419} {From system 1 to
  system 2: {A} survey of reasoning large language models}.
\newblock \emph{CoRR}, abs/2502.17419.

\bibitem[{Liu et~al.(2025)Liu, Lyu, Wu, Wang, Luo, Zhang, and
  Shang}]{Newtrends}
Sinuo Liu, Chenyang Lyu, Minghao Wu, Longyue Wang, Weihua Luo, Kaifu Zhang, and
  Zifu Shang. 2025.
\newblock \href {https://doi.org/10.48550/ARXIV.2503.10351} {New trends for
  modern machine translation with large reasoning models}.
\newblock \emph{CoRR}, abs/2503.10351.

\bibitem[{Luo et~al.(2025)Luo, Tan, Wong, Shi, Tang, Roongta, Cai, Luo, Zhang,
  Li, Popa, and Stoica}]{deepscaler2025}
Michael Luo, Sijun Tan, Justin Wong, Xiaoxiang Shi, William~Y. Tang, Manan
  Roongta, Colin Cai, Jeffrey Luo, Tianjun Zhang, Li~Erran Li, Raluca~Ada Popa,
  and Ion Stoica. 2025.
\newblock Deepscaler: Surpassing o1-preview with a 1.5b model by scaling rl.
\newblock \url{https://github.com/agentica-project/deepscaler}.
\newblock Notion Blog.

\bibitem[{Moslem et~al.(2023)Moslem, Romani, Molaei, Kelleher, Haque, and
  Way}]{term2}
Yasmin Moslem, Gianfranco Romani, Mahdi Molaei, John~D. Kelleher, Rejwanul
  Haque, and Andy Way. 2023.
\newblock \href {https://doi.org/10.18653/V1/2023.WMT-1.82} {Domain terminology
  integration into machine translation: Leveraging large language models}.
\newblock In \emph{Proceedings of the Eighth Conference on Machine Translation,
  {WMT} 2023, Singapore, December 6-7, 2023}, pages 902--911. Association for
  Computational Linguistics.

\bibitem[{Myung et~al.(2024)Myung, Park, Son, Lee, and Han}]{term3}
Jiyoon Myung, Jihyeon Park, Jungki Son, Kyungro Lee, and Joohyung Han. 2024.
\newblock \href {https://doi.org/10.18653/v1/2024.wmt-1.129} {Efficient
  technical term translation: A knowledge distillation approach for
  parenthetical terminology translation}.
\newblock In \emph{Proceedings of the Ninth Conference on Machine Translation},
  pages 1410--1427, Miami, Florida, USA. Association for Computational
  Linguistics.

\bibitem[{Papineni et~al.(2002)Papineni, Roukos, Ward, and Zhu}]{bleu}
Kishore Papineni, Salim Roukos, Todd Ward, and Wei{-}Jing Zhu. 2002.
\newblock \href {https://doi.org/10.3115/1073083.1073135} {Bleu: a method for
  automatic evaluation of machine translation}.
\newblock In \emph{Proceedings of the 40th Annual Meeting of the Association
  for Computational Linguistics, July 6-12, 2002, Philadelphia, PA, {USA}},
  pages 311--318. {ACL}.

\bibitem[{Post(2018)}]{bleu2}
Matt Post. 2018.
\newblock \href {https://doi.org/10.18653/V1/W18-6319} {A call for clarity in
  reporting {BLEU} scores}.
\newblock In \emph{Proceedings of the Third Conference on Machine Translation:
  Research Papers, {WMT} 2018, Belgium, Brussels, October 31 - November 1,
  2018}, pages 186--191. Association for Computational Linguistics.

\bibitem[{Qin et~al.(2024)Qin, Li, Zou, Liu, Xia, Huang, Ye, Yuan, Liu, Li, and
  Liu}]{o1replicationjourney}
Yiwei Qin, Xuefeng Li, Haoyang Zou, Yixiu Liu, Shijie Xia, Zhen Huang, Yixin
  Ye, Weizhe Yuan, Hector Liu, Yuanzhi Li, and Pengfei Liu. 2024.
\newblock \href {https://doi.org/10.48550/ARXIV.2410.18982} {{O1} replication
  journey: {A} strategic progress report - part 1}.
\newblock \emph{CoRR}, abs/2410.18982.

\bibitem[{Rei et~al.(2020)Rei, Stewart, Farinha, and Lavie}]{comet22}
Ricardo Rei, Craig Stewart, Ana~C. Farinha, and Alon Lavie. 2020.
\newblock \href {https://doi.org/10.18653/V1/2020.EMNLP-MAIN.213} {{COMET:} {A}
  neural framework for {MT} evaluation}.
\newblock In \emph{Proceedings of the 2020 Conference on Empirical Methods in
  Natural Language Processing, {EMNLP} 2020, Online, November 16-20, 2020},
  pages 2685--2702. Association for Computational Linguistics.

\bibitem[{Rei et~al.(2022)Rei, Treviso, Guerreiro, Zerva, Farinha, Maroti,
  de~Souza, Glushkova, Alves, Coheur, Lavie, and Martins}]{cometkiwi}
Ricardo Rei, Marcos~V. Treviso, Nuno~Miguel Guerreiro, Chrysoula Zerva, Ana~C.
  Farinha, Christine Maroti, Jos{\'{e}} G.~C. de~Souza, Taisiya Glushkova,
  Duarte~M. Alves, Lu{\'{\i}}sa Coheur, Alon Lavie, and Andr{\'{e}} F.~T.
  Martins. 2022.
\newblock \href {https://aclanthology.org/2022.wmt-1.60} {Cometkiwi:
  Ist-unbabel 2022 submission for the quality estimation shared task}.
\newblock In \emph{Proceedings of the Seventh Conference on Machine
  Translation, {WMT} 2022, Abu Dhabi, United Arab Emirates (Hybrid), December
  7-8, 2022}, pages 634--645. Association for Computational Linguistics.

\bibitem[{Sabet et~al.(2020)Sabet, Dufter, Yvon, and Sch{\"{u}}tze}]{simalign}
Masoud~Jalili Sabet, Philipp Dufter, Fran{\c{c}}ois Yvon, and Hinrich
  Sch{\"{u}}tze. 2020.
\newblock \href {https://doi.org/10.18653/V1/2020.FINDINGS-EMNLP.147}
  {Simalign: High quality word alignments without parallel training data using
  static and contextualized embeddings}.
\newblock In \emph{Findings of the Association for Computational Linguistics:
  {EMNLP} 2020, Online Event, 16-20 November 2020}, volume {EMNLP} 2020 of
  \emph{Findings of {ACL}}, pages 1627--1643. Association for Computational
  Linguistics.

\bibitem[{Schulman et~al.(2017)Schulman, Wolski, Dhariwal, Radford, and
  Klimov}]{ppo}
John Schulman, Filip Wolski, Prafulla Dhariwal, Alec Radford, and Oleg Klimov.
  2017.
\newblock \href {https://arxiv.org/abs/1707.06347} {Proximal policy
  optimization algorithms}.
\newblock \emph{Preprint}, arXiv:1707.06347.

\bibitem[{Sellam et~al.(2020)Sellam, Das, and Parikh}]{bleurt}
Thibault Sellam, Dipanjan Das, and Ankur Parikh. 2020.
\newblock \href {https://doi.org/10.18653/v1/2020.acl-main.704} {{BLEURT}:
  Learning robust metrics for text generation}.
\newblock In \emph{Proceedings of the 58th Annual Meeting of the Association
  for Computational Linguistics}, pages 7881--7892, Online. Association for
  Computational Linguistics.

\bibitem[{Shao et~al.(2024)Shao, Wang, Zhu, Xu, Song, Bi, Zhang, Zhang, Li, Wu,
  and Guo}]{ds-math}
Zhihong Shao, Peiyi Wang, Qihao Zhu, Runxin Xu, Junxiao Song, Xiao Bi, Haowei
  Zhang, Mingchuan Zhang, Y.~K. Li, Y.~Wu, and Daya Guo. 2024.
\newblock \href {https://arxiv.org/abs/2402.03300} {Deepseekmath: Pushing the
  limits of mathematical reasoning in open language models}.
\newblock \emph{Preprint}, arXiv:2402.03300.

\bibitem[{Song et~al.(2025)Song, Zheng, Li, Yang, Luo, Pan, and
  Zhang}]{fastcurl}
Mingyang Song, Mao Zheng, Zheng Li, Wenjie Yang, Xuan Luo, Yue Pan, and Feng
  Zhang. 2025.
\newblock \href {https://arxiv.org/abs/2503.17287} {Fastcurl: Curriculum
  reinforcement learning with progressive context extension for efficient
  training r1-like reasoning models}.
\newblock \emph{Preprint}, arXiv:2503.17287.

\bibitem[{Wang et~al.(2024)Wang, Meng, Liang, and Zhou}]{DRT}
Jiaan Wang, Fandong Meng, Yunlong Liang, and Jie Zhou. 2024.
\newblock \href {https://doi.org/10.48550/ARXIV.2412.17498} {Drt-o1: Optimized
  deep reasoning translation via long chain-of-thought}.
\newblock \emph{CoRR}, abs/2412.17498.

\bibitem[{Wang et~al.(2025)Wang, Meng, and Zhou}]{Deeptrans}
Jiaan Wang, Fandong Meng, and Jie Zhou. 2025.
\newblock \href {https://arxiv.org/abs/2504.10187} {Deep reasoning translation
  via reinforcement learning}.
\newblock \emph{Preprint}, arXiv:2504.10187.

\bibitem[{Yang et~al.(2024)Yang, Yang, Zhang, Hui, Zheng, Yu, Li, Liu, Huang,
  Wei, Lin, Yang, Tu, Zhang, Yang, Yang, Zhou, Lin, Dang, Lu, Bao, Yang, Yu,
  Li, Xue, Zhang, Zhu, Men, Lin, Li, Xia, Ren, Ren, Fan, Su, Zhang, Wan, Liu,
  Cui, Zhang, and Qiu}]{Qwen2.5}
An~Yang, Baosong Yang, Beichen Zhang, Binyuan Hui, Bo~Zheng, Bowen Yu,
  Chengyuan Li, Dayiheng Liu, Fei Huang, Haoran Wei, Huan Lin, Jian Yang,
  Jianhong Tu, Jianwei Zhang, Jianxin Yang, Jiaxi Yang, Jingren Zhou, Junyang
  Lin, Kai Dang, and 22 others. 2024.
\newblock \href {https://doi.org/10.48550/ARXIV.2412.15115} {Qwen2.5 technical
  report}.
\newblock \emph{CoRR}, abs/2412.15115.

\bibitem[{Zeng et~al.(2025)Zeng, Huang, Liu, He, Liu, Ma, and
  He}]{zeng2025simplerl}
Weihao Zeng, Yuzhen Huang, Wei Liu, Keqing He, Qian Liu, Zejun Ma, and Junxian
  He. 2025.
\newblock 7b model and 8k examples: Emerging reasoning with reinforcement
  learning is both effective and efficient.
\newblock \url{https://hkust-nlp.notion.site/simplerl-reason}.
\newblock Notion Blog.

\bibitem[{Zhang et~al.(2023)Zhang, Wang, Qin, Shi, Wang, and Chen}]{rtt}
Huaao Zhang, Qiang Wang, Bo~Qin, Zelin Shi, Haibo Wang, and Ming Chen. 2023.
\newblock \href {https://doi.org/10.18653/V1/2023.ACL-LONG.332} {Understanding
  and improving the robustness of terminology constraints in neural machine
  translation}.
\newblock In \emph{Proceedings of the 61st Annual Meeting of the Association
  for Computational Linguistics (Volume 1: Long Papers), {ACL} 2023, Toronto,
  Canada, July 9-14, 2023}, pages 6029--6042. Association for Computational
  Linguistics.

\bibitem[{Zhang et~al.(2024)Zhang, Wu, Yang, Shu, Xiao, Kong, and
  Sang}]{o1-coder}
Yuxiang Zhang, Shangxi Wu, Yuqi Yang, Jiangming Shu, Jinlin Xiao, Chao Kong,
  and Jitao Sang. 2024.
\newblock \href {https://doi.org/10.48550/ARXIV.2412.00154} {o1-coder: an o1
  replication for coding}.
\newblock \emph{CoRR}, abs/2412.00154.

\bibitem[{Zhao et~al.(2024)Zhao, Yin, Zeng, Wang, Shi, Lyu, Wang, Luo, and
  Zhang}]{marco-o1}
Yu~Zhao, Huifeng Yin, Bo~Zeng, Hao Wang, Tianqi Shi, Chenyang Lyu, Longyue
  Wang, Weihua Luo, and Kaifu Zhang. 2024.
\newblock \href {https://doi.org/10.48550/ARXIV.2411.14405} {Marco-o1: Towards
  open reasoning models for open-ended solutions}.
\newblock \emph{CoRR}, abs/2411.14405.

\bibitem[{Zheng et~al.(2024)Zheng, Hong, Liu, Wang, Su, Liang, and Wu}]{dragft}
Jiawei Zheng, Hanghai Hong, Feiyan Liu, Xiaoli Wang, Jingsong Su, Yonggui
  Liang, and Shikai Wu. 2024.
\newblock \href {https://arxiv.org/abs/2402.15061} {Fine-tuning large language
  models for domain-specific machine translation}.
\newblock \emph{Preprint}, arXiv:2402.15061.

\end{thebibliography}

\appendix

\end{document}